\pdfoutput=1

\documentclass[11pt]{article}

\usepackage[]{EMNLP2022}

\usepackage{times}
\usepackage{latexsym}
\usepackage{graphicx}

\usepackage[T1]{fontenc}

\usepackage[utf8]{inputenc}

\usepackage{microtype}

\usepackage{inconsolata}

\usepackage{multirow}

\title{Entity-level Sentiment Analysis in Contact Center Telephone Conversations}

\author{Xue-Yong Fu, \ Cheng Chen, \ Md Tahmid Rahman Laskar, \\ {\bf Shayna Gardiner, Pooja Hiranandani, Shashi Bhushan TN}\\
        Dialpad Canada Inc.\\ Vancouver, BC, Canada\\
        \texttt{\{xue-yong,cchen,tahmid.rahman\}@dialpad.com}\\\texttt{\{sgardiner,phiranandani,sbhushan\}@dialpad.com}}

\begin{document}
\maketitle
\begin{abstract}

Entity-level sentiment analysis predicts the sentiment about entities mentioned in a given text. It is very useful in a business context to understand user emotions towards certain entities, such as products or companies. In this paper, we demonstrate how we developed an entity-level sentiment analysis system that analyzes English telephone conversation transcripts in contact centers to provide business insight. We present two approaches, one entirely based on the transformer-based DistilBERT model, and another that uses a convolutional neural network supplemented with some heuristic rules.

\end{abstract}

\section{Introduction}

Businesses that provide Contact Center as a Service (CCaaS) often leverage Artificial Intelligence (AI) technologies to transcribe telephone conversations and generate aggregated insights reports for contact centers across various industry verticals. 
Customers occasionally make evaluative comments about specific products or companies during a customer support call to a contact center. These comments provide valuable competitive insights to the business, e.g. positive comments may provide information useful to a marketing department as it formulates an advertising campaign while negative comments may provide valuable insights that can be used to improve a product or a service. In such scenarios, building a system that can identify user sentiments towards entities like product or companies could be useful.

Though Aspect-Based Sentiment Analysis (ABSA) \cite{zhou2019deep, DBLP:journals/kbs/ZhouHHH20} that aims to extract the user sentiment expressed towards a specific aspect associated with a given target could be a possible solution to solve such problems, it should be noted that ABSA has a key limitation in this regard. For example, in tasks like competitor analysis where the objective is to understand the overall user sentiment within a certain period of time on specific products or general company services, ABSA techniques could not be useful as they provide more fine-grained opinions towards predefined aspects (e.g. features of a product, ease of use of a software, or aspects of restaurant experience) instead of providing a generic user sentiment towards a specific entity \cite{zhou2019deep}. 

While building the Entity-level Sentiment Analysis (ELSA) system for real-world contact center use-cases, we observe several key challenges. First of all, to the best of our knowledge, there are no existing public datasets available for the entity-level sentiment analysis task. Meanwhile, this task becomes more challenging when the requirement is to use a dataset constructing from telephone conversations since constructing a dataset from speech transcripts generated from telephone conversations is non-trivial as telephone transcripts are generated by automatic speech recognition (ASR) systems that have their own unique characteristics. For instance, an ASR system may have mistranscription errors as well as linguistic disfluencies (e.g. filler words) \cite{fu2021improving} that usually occur in a conversational speech dataset \cite{malik2021automatic}.

The factors mentioned above make the implementation of an entity-level sentiment analysis model very challenging to detect user opinions towards entities that appear in contact center calls. In this paper, we address the existing limitations behind developing an entity-sentiment model for commercial  scenarios in the domain of business telephone conversation data in contact centers. Since there is no suitable publicly available dataset for the entity-level sentiment analysis task, we briefly describe how we sampled and annotated the data for this task. We then propose two approaches that leverage neural models for this task (i) one is based on the DistilBERT~\cite{Sanh2019DistilBERTAD} model in which we modify its architecture such that the model can also be utilized to extract the opinion term(s) while detecting the sentiment polarity towards a named entity; (ii) while the other approach uses a convolutional neural network \cite{DBLP:conf/coling/SantosG14} supplemented with some pre-defined heuristic rules. We compare the effectiveness of both approaches through extensive experiments and discuss our findings to provide valuable insights for future developments of ELSA models for real world commercial scenarios.



\section{Related Work}

Since the entity-level sentiment analysis task is closely related to aspect-level sentiment analysis, in this section, we first briefly review the aspect-level sentiment analysis task followed by the entity-level sentiment analysis task in order to clarify the distinction between these two tasks while discussing our rationale behind developing an entity-level sentiment analysis model for contact centers.

\subsection{Aspect-Based Sentiment Analysis (ABSA)}
ABSA aims to classify the sentiment polarity of aspects of certain objects. Many previous studies are focused on this research~\cite{DBLP:conf/naacl/SunHQ19, DBLP:conf/emnlp/TangQL16, DBLP:conf/coling/HeLND18, zhao-etal-2020-attention, zhou2019deep}. A more fine-grained related task is aspect sentiment triplet extraction (ASTE)~\cite{DBLP:conf/aaai/PengXBHLS20, DBLP:conf/emnlp/XuLLB20}, which extracts a triplet -- aspect term, opinion term and sentiment -- from the input. Detection of aspects in both ABSA or ASTE often relies on implicit lexical or semantic signs, for instance, \textit{the food is too spicy} suggests that this comment is about the \textit{taste} aspect. This is different from the entity recognition task where the goal is to detect the named entities in a given utterance based on the overall context. 



\subsection{Entity-level Sentiment Analysis (ELSA)}
ELSA aims to predict the sentiment of named entities in a given text input \cite{steinberger2011multilingual,saif2014semantic}. These named entities are usually application dependent. One recent work on ELSA is the work of~\citet{luo2022entity}, where they studied entity sentiment in news documents. Another prominent work on ELSA is the work of ~\citet{DBLP:conf/icse/Ding00L18}, where an entity-level sentiment analysis tool was proposed for Github issue comments. 
Contrary to the above studies that focused on typed text, our focus is on noisy textual data (i.e., speech transcripts).  
Moreover, our proposed models can infer both entity sentiment and corresponding opinion terms for a better analysis of user sentiments towards products or companies in business telephone conversations in contact centers.






\section{Task Description}

Let us assume that we have an utterance $U = w_1, w_2,...,w_n$ containing $n$ words. The goal of the ELSA task is to identify $m$ opinion words $O_W = ow_1, ow_2,...,ow_m$, (where $m$ < $n$), and classify the sentiment of the identified opinion words towards the target entity $e$ in the given utterance. In Table~\ref{tab:example}, we show some examples of the ELSA task to detect user sentiments towards products and organization type entities. 
In the first two examples, the customer is directly expressing positive sentiment about the named entity. For instance, (i) they say “I love it” indicating “Google” in context, or (ii) they are “very impressed” with “MAC”. In the third and fourth examples, customers are expressing negative sentiment about a product or facet associated with the company, e.g., “He has a hard time finding a good yogurt from Walmart” is a comment about the quality of Walmart’s service, not a comment about yogurt. Similarly, in the fourth example, difficulty navigating the Instacart app is indirectly an indication of negative sentiment concerning Instacart.

\begin{table*}[]
\centering
\begin{tabular}{|c|}

\hline
I work at \textcolor{blue}{Google} and I \textcolor{teal}{love} it a lot.                                \\ \hline
She's \textcolor{teal}{very impressed} how \textcolor{blue}{MAC} works so well.                      \\ \hline
He has \textcolor{purple}{hard time} finding a good yogurt from \textcolor{blue}{Walmart}.                  \\ \hline
It's \textcolor{purple}{quite difficult} to navigate the mobile app of \textcolor{blue}{Instacart}.           \\ \hline
\end{tabular}
\caption{Examples for entity-level sentiment analysis. Words in color \textit{\textcolor{blue}{blue}} are target named entities. Words in color \textit{\textcolor{teal}{teal}} are positive opinion words. Words in color  \textit{\textcolor{purple}{purple}} are negative opinion words. }
\label{tab:example}

\end{table*}


\section{Dataset Construction}

As noted earlier, there is no publicly available dataset for the ELSA task. We therefore had to create and annotate our own dataset. The first major issue that we observed while constructing a dataset for ELSA is that the entity-level sentiment events in our telephone transcripts are very infrequent. Hence, random data sampling techniques might yield an imbalanced dataset where most utterances would not have any positive or negative sentiments towards an entity. We therefore used two pre-existing models — a named entity recognition (NER) model based on DistilBERT \cite{Sanh2019DistilBERTAD} that was trained to identify \textbf{Organization} and \textbf{Product} type entities and a convolutional neural network (CNN) ~\cite{DBLP:conf/nips/KrizhevskySH12, DBLP:conf/coling/SantosG14, albawi2017understandingcnn} sentiment analysis model — to sample 13000 utterances that contained at least one named entity and one positive or negative sentiment predicted by these models. To balance the dataset, we sampled an additional 10000 utterances containing at least one entity and having no polarized sentiments (i.e., only neutral sentiment). The resulting 23000 utterances were manually annotated by independent annotators to determine the positive, neutral, or negative sentiment toward the target entity. The annotators also identified the opinion terms in the utterances.

\section{Our Proposed Models}

For performance evaluation, we propose two approaches: \textit{(i) DistilBERT-based Model}, and \textit{(ii) CNN-based Model with Heuristics Rules}. Below, we present our proposed approaches. 

\subsection{DistilBERT-based Model}

For this approach, we leverage the DistilBERT model since this is a very lightweight model that does not require much computing power in production environments \cite{Sanh2019DistilBERTAD}. Below, we describe how we utilize this model for ELSA.

\paragraph{NER tagging:}
Given an utterance as input, we first run an NER model to determine if there is at least one entity (product or organization) detected. Our NER model is based on DistilBERT that is trained over business conversation data collected from call centers. During the training stage, we use the cross entropy (CE) loss as defined in Equation~\ref{eqn:cross_entropy}:
\begin{equation}
\label{eqn:cross_entropy}
L_{CE} = - \frac{1}{N} \sum_{n=1}^{N}\log\frac{e^{\hat{y}_{n, y_n}}}{\sum_{c=1}^{C}e^{\hat{y}_{n,c}}}
\end{equation}
Here, $N$ is the number of samples in a batch, and $C$ denotes the number of classes, $\hat{y}_{n, c}$ is the logit of the $c$-th class in the $n$-th example, and $\hat{y}_{n, y_n}$ is the logit of the gold class in the $n$-th example.

\paragraph{Context Representation:}
We insert a special tag, \textbf{\textit{ \_NE\_}}, before any named entities detected in an utterance. This helps the model to identify which spans belong to the entity. For example, if the raw input is “\textit{I really don’t like using Snapchat}”, we reformulated the input as "\textit{I really don’t like using \_NE\_ Snapchat}". Then, we send the pre-processed input to our entity sentiment detection model that we describe below.

\paragraph{Entity Sentiment Detection:}

Our entity sentiment detection model is also based on DistilBERT. However, for this task, we train DistilBERT over business telephone conversation data for a different task: the sentiment classification task. Meanwhile, our entity sentiment detection model can also extract the opinion word(s) in a given utterance. This is done by adding an additional prediction layer on top of the DistilBERT model to identify the opinion words. During the training phase, the model is fine-tuned on our entity sentiment dataset to predict the polarity of the opinion terms for a given utterance. If the target entity's sentiment is positive or negative, the model will assign respective tags (\textbf{POS} for positive and \textbf{NEG} for negative)  to the opinion token(s), while the remaining tokens will be assigned to the \textbf{O} tag.



\textit{\textbf{Transfer Learning:}} To improve model performance, we introduce a transfer learning technique for our entity sentiment detection model, for which we first fine-tune the DistilBERT model for the sentence classification task (i.e., sentiment analysis) on the Stanford Sentiment Treebank (SST) dataset that contains $11,855$ training examples. The assumption is that if a model is fine-tuned on a similar task, it is expected to perform better on related downstream tasks \cite{laskar2022domain,garg2020tanda}. Although the SST dataset is about predicting the general sentiment of a given text sequence, it requires the model to learn what words are associated with positive sentiments and what are associated with negative sentiment, which is essential for our task. The DistilBERT model trained on the SST dataset is then fine-tuned again on the processed input (pre-processed by using the \textbf{\textit{ \_NE\_}} tag obtained from our NER model) in the contact center conversation dataset. As mentioned earlier, in this stage of fine-tuning, the entity sentiment model can also extract the opinion word(s) via utilizing the additional prediction layer that we added on top of DistilBERT. 

\begin{figure*}[t!]
\begin{center}
\includegraphics[width=\linewidth]{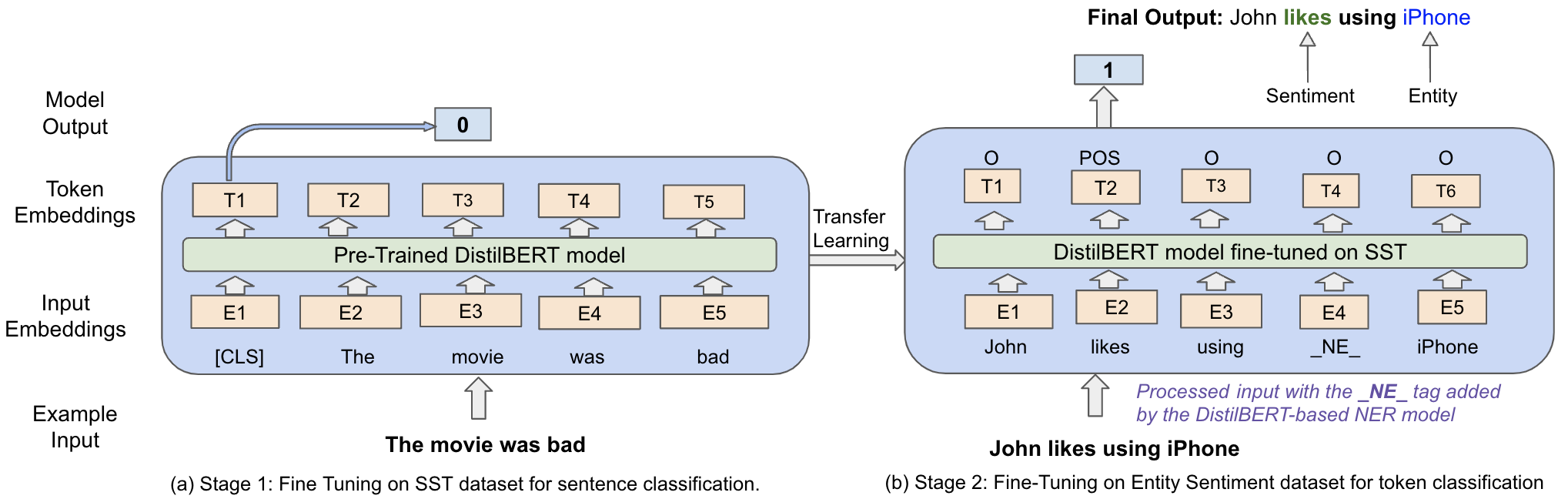}
\caption{An overview of our proposed DistilBERT-based approach: (a) first, we do fine-tuning on the SST dataset for the generic sentiment analysis task, and (b) then, fine-tune on the in-domain Entity Sentiment dataset for entity level opinion extraction. Here, in the output layer, 1 denotes positive while 0 denotes negative sentiment. }
\label{fig:ModelOverview}
\end{center}
\end{figure*}

An overview of our proposed DistilBERT-based approach is shown in Figure \ref{fig:ModelOverview}.









\subsection{CNN-based Sentiment Model Supplemented with Heuristic Rules}
For this model, we employ a two-step approach. We first run a general sentiment analysis model that classifies the sentiment of a given utterance and also extracts the keywords that cause that sentiment (if the sentiment is positive or negative). We treat these sentiment keywords as opinion word candidates. Then we employ a set of linguistic heuristics that identify the opinion words that are associated with the entities mentioned in the input.

The sentiment analysis model is a multiclass, CNN-based classification model. We choose CNN here due to its effectiveness in related tasks (e.g., ABSA task) \cite{wang2021unified}. However, for ELSA, we add an explainability layer on top of CNN that is tasked with explaining predictions. The explainability technique that we leverage is called Integrated Gradients, adopted from ~\cite{DBLP:conf/icml/SundararajanTY17}. After a sentiment score is predicted by the model, the explainability layer emits words that are highly associated with the predicted sentiment. Note that we apply some heuristics to select the emitted words as candidates for opinion words.

\paragraph{\textbf{Heuristics:}} 
For the extraction of opinion words, we utilize heuristics based on phrase structure types that are most likely to contain entity sentiment. We divide these phrase types into three categories to find which part of speech contains the potential opinion word: verb-based, adjective-based, and noun-based. Table \ref{tab:ling_heuristics} illustrates the possible syntactic patterns captured by these heuristics, all of which also allow for optional modifiers such as intensifiers (e.g. \textit{really, very, so}), complementizers (\textit{that, which}) or stacked adjectives. Some example phrases that were captured include: \textit{I’m so happy that Google made this}, \textit{Android sucks}, \textit{that was awesome of Netflix to do}, \textit{Netflix is garbage}, \textit{my hatred of LaTeX}, \textit{classic LaTeX awesomeness}, etc.

\begin{table*}[tbh]
\centering
\small
\begin{tabular}{|c|c|c|}
\hline
\textbf{Verb-Based} & \textbf{Adjective-Based} & \textbf{Noun-Based}\\ \hline
sentiment verb & sentiment adjective & entity\\ 
+ entity & + entity & + sentiment noun\\
\hline
sentiment verb + & sentiment adjective & sentiment noun\\
+ preposition &  + \textit{of}/\textit{for} + entity & preposition \\
+ entity & & + entity \\
\hline
entity + & & entity + aux verb \\ 
sentiment verb & & sentiment noun \\ \hline
\end{tabular}
\caption{Heuristic Rules to Extract Opinion Words}
\label{tab:ling_heuristics}
\end{table*}

\section{Experiment}

In this section, we present the training parameters, evaluation metrics, and the experimental results. In our experiments, we use the following three models:

\begin{itemize}
    \item \textbf{DistilBERT}: This model does not leverage the SST dataset as the first stage of fine-tuning. Instead, it is fine-tuned only on our ELSA dataset. 
    \item \textbf{DistilBERT + SST}: This model is initially fine-tuned on the SST dataset and then fine-tuned on our ELSA dataset.
    \item \textbf{CNN + Heuristics}: This is the model that leverages CNN and supplemented with some heuristics rules. 
\end{itemize}

\subsection{Training Parameters}
For the DistilBERT model, we set the batch size to 32, learning rate to $5\times10^{-5}$, and employ early stopping with patience set to 5. The pretrained model is based on the HuggingFace Transformer \cite{DBLP:conf/emnlp/WolfDSCDMCRLFDS20}. While for the CNN model, we use 300 dimensional fastText embeddings \cite{bojanowski2016enriching,santos2017sentiment}, global max pooling is utilized in the convolational layer with filter sizes: 2, 3, 4, 5, 6. The fully connected layer is 128 dimensional. 


\begin{table*}[t]
\centering
\small
\begin{tabular}{|cc|c|c|c|c|c|c|}
\hline
\multicolumn{2}{|c|}{\multirow{2}{*}{\textbf{Models}}} &\multicolumn{2}{c|}{{\textbf{Precision}}}  & \multicolumn{2}{c|}{{\textbf{ Recall}}}     & \multicolumn{2}{c|}{{\textbf{F1}}}     
\\ \cline{3-8}
\multicolumn{2}{|c|}{}& \textbf{Ent} & \textbf{OP} & \textbf{Ent} & \textbf{Op} & \textbf{Ent} & \textbf{Op}  
\\ \cline{1-8}

\multicolumn{2}{|c|}{\textbf{DistilBERT}}& 74.43 & 59.99 & 73.77 & 69.44 & 73.70 & 64.35 \\ \hline
\multicolumn{2}{|c|}{\textbf{DistilBERT + SST}}& 74.83 & 68.21 & 74.69 & 63.02 & 74.72 & 65.48 \\ \hline

\multicolumn{2}{|c|}{\textbf{CNN + Heuristics}}& 77.48 & 97.65 & 58.23 & 16.78 & 50.07 & 28.64 \\ \hline
\end{tabular}
\caption{Experimental results for the Entity Level Sentiment (Ent) and Opinion Word Extraction (OP) tasks. }
\label{tab:ent_op_val} 
\end{table*}

\subsection{Evaluation Metrics}
To evaluate entity sentiment classification and opinion word extraction, we define two kinds of evaluation metric. For polarity classification, we calculate precision, recall and F1 score for three sentiment categories: positive, negative and neutral. We then calculate the weighted value of these (support-based). For opinion word extraction, we evaluate it using the metrics that are usually used for named entity recognition \cite{li2020survey} and calculate precision, recall and F1 score for opinion words. The evaluation was done in a sample of 175 annotated utterances that were reviewed by another group of annotators.


\subsection{Results}

From Table \ref{tab:ent_op_val}, we find that in terms of the F1 metric, both  variations of DistilBERT -- (Vanilla \textbf{DistilBERT} and \textbf{DistilBERT + SST}) -- outperform the \textbf{CNN + Heuristics}  model by a huge margin. More specifically, \textbf{DistilBERT + SST} outperforms the \textbf{CNN + Heuristics model} by $15.75\%$ of the F1 score. Comparing \textbf{DistilBERT} and \textbf{DistilBERT + SST}, we can see the effect of SST pre-training, which brings the F1 score up from $73.7\%$ to $74.72\%$, with an increase of 1.38\%. We also find that the \textbf{CNN + Heuristics} model obtains impressive precision score. This is because the heuristic rules used in the \textbf{CNN + Heuristics} model were developed to emphasize precision, but they do not handle linguistic variation well, resulting in poor Recall and F1 scores.

For opinion word extraction, which is noted as \textbf{OP}, the performance gap between the \textbf{DistilBERT} model and the \textbf{CNN + Heuristics} model is even larger. As shown in Table \ref{tab:ent_op_val}, \textbf{DistilBERT + SST} outperforms the \textbf{CNN + Heuristics} by 38.48\% F1 score.  This is mainly because the \textbf{CNN + Heuristics} has very poor performance in recall: only 16\%. Although the recall of \textbf{DistilBERT + SST} is lower than \textbf{DistilBERT}, its F1 score is still 1.07\% higher than its counterpart.  

\paragraph{Robustness Test:} The overall metrics can’t identify if the performance of a model is robust in different situations. Thus, we investigate if our proposed model is robust against various kinds of input texts. For this purpose, we separate the test data into different sub-populations by the number of tokens and the number of entities. We then evaluate our models on sub-populations to see how they perform. Below, we define these sub-populations.  

\begin{table*}[]
\small
\begin{center}
\begin{tabular}{|c|c|c|c|}
\hline
\textbf{Model}                                             & \textbf{Precision} & \textbf{Recall} & \textbf{F1} \\ \hline
\textbf{CNN + Heuristics}                                          & 97.65              & 16.78           & 28.64       \\ \hline
\textbf{CNN + Heuristics (\textless{} 8 tokens)}                     & 100                & 38.89           & 56          \\ \hline
\textbf{CNN + Heuristics (\textgreater{} 45 tokens)}                 & 100                & 8.16            & 14.99       \\ \hline
\textbf{CNN + Heuristics (= 1 entity)}                     & 97.54              & 16.86           & 28.75       \\ \hline
\textbf{CNN + Heuristics (\textgreater{} 1 entity)}          & 100                & 15.56           & 26.72       \\ \hline
\textbf{DistilBERT + SST}                      & 68.21              & 63.02           & 65.48       \\ \hline
\textbf{DistilBERT + SST (\textless{} 8 tokens)} & 66.9               & 79.46           & 72.46       \\ \hline
\textbf{DistilBERT + SST (\textgreater{} 45 tokens) }                    & 62.18              & 26.32           & 36.92       \\ \hline
\textbf{DistilBERT + SST (= 1 entity)} & 68.28              & 63.67           & 65.87       \\ \hline
\textbf{DistilBERT + SST (\textgreater{} 1 entity)}             & 68                 & 52.94           & 58.71       \\ \hline
\end{tabular}
\end{center}
\caption{Robustness Report on the Opinion Word Extraction task.}
\label{tab:ent_op_val_slices}
\end{table*}


\textit{\textbf{(i) 1 entity:} input text has only one target entity.}

\textit{\textbf{(ii) > 1 entity:} input text has more than one target entity.} 

\textit{\textbf{(iii) < 8 tokens:} input text with less than eight tokens.} 

\textit{\textbf{(iv) > 45 tokens:} input with more than forty five tokens.} 

Table~\ref{tab:ent_op_val_slices} contains the results of our proposed models in different data slices. We find that both models perform poorly when the input is long (> 46 tokens) compared to when the input is short (< 8 tokens). This could be because it is much harder to model long term dependencies when the sequence length is too long. 

We also find that the \textbf{DistilBERT + SST} model is more sensitive to the number of target entities in the input compared to the \textbf{CNN + Heuristics} model. Its F1 score drops by $7.16\%$ when the number of target entities increases from one to more than one. 

\section{Commercial Application}

We have deployed the \textbf{DistilBERT + SST} model in our production system to generate entity sentiment data for contact centers as it has better accuracy than the \textbf{CNN + Heuristics} model. Due to the small model size and efficient inference of DistilBERT, each model instance is assigned 1 CPU and 1GB memory. Once there are enough entity-level sentiment predictions, there is a dedicated pipeline to aggregate entity sentiment in different granularity for each customer. 

There are many use cases of the aggregated insights of entity sentiment. Contact center managers can use this information to improve contact center efficiency by investigating why customers are not happy with certain products (e.g., \textit{itelephone 13 Pro Max}) and develop desired responses when the customer is complaining about it, so that the agents can handle the difficult situation more efficiently. The collected negative feedback can be used to inform the product team how to improve the products. The insights can also be used to conduct comparisons between several products or companies and help with competitor analysis.






\section{Conclusion}
In this paper, we described the creation of a task-specific dataset and a new model that extracts opinion words while performing entity sentiment polarity detection. The resulting DistilBERT-based model is currently deployed as a commercial application for entity-level sentiment analysis for English contact center conversations. In the future, we will investigate how to extend our proposed methods to other applications \cite{laskar2022auto, laskar-etal-2022-blink} of the entity recognition task \cite{funercoling} in telephone transcripts and explore how to improve model performance on utterances that contain more than one entity.

\section*{Limitations}
As our entity sentiment models are trained on English business telephone conversations, they might not be suitable to be used in other domains, types of inputs (i.e written text), or languages. The NER component of \textbf{DistilBERT} based model has some limitations while detecting product and organization type entities. It is more biased towards detecting the entities that appear more frequently in the training data and misses rare entities. This could impact the overall performance of the model.




\section*{Ethics Statement}
This data in this research is comprised of individual sentences that do not contain sensitive, personal, or identifying information.  The entity sentiment model deployed in production is not used to attach any sentiment to people, only to non-human entities. Each machine-sampled utterance is labelled by annotators before the utterance is used as part of the training dataset. While annotator demographics are unknown and therefore may introduce potential bias in the labelled dataset, the annotators are required to pass a screening test before completing any labels used in these experiments, thereby mitigating this unknown to some extent. We paid adequate compensation to the annotators. Future work should nonetheless strive to improve training data further in this regard.

\section*{ACKNOWLEDGEMENTS}
We would like to thank Simon Corston-Oliver for his helpful and detailed feedback on the paper. We also thank the reviewers for their excellent review comments that helped us improving the paper.


\bibliography{anthology,custom}
\bibliographystyle{acl_natbib}



\end{document}